\documentclass{ecai} 

\usepackage{latexsym}
\usepackage{amssymb}
\usepackage{amsmath}
\usepackage{amsthm}
\usepackage{booktabs}
\usepackage{enumitem}
\usepackage{graphicx}
\usepackage{color}
\usepackage{multirow}
\usepackage{graphicx}
\usepackage{caption}
\usepackage{xcolor}

\newcommand{\BibTeX}{B\kern-.05em{\sc i\kern-.025em b}\kern-.08em\TeX}
\newcommand{\mysmall}[1]{\scriptsize{\color{gray}{#1}}}

\begin{document}


\begin{frontmatter}


\paperid{6676} 


\title{D$^3$-Talker: Dual-Branch Decoupled Deformation Fields for Few-Shot 3D Talking Head Synthesis}


\author[A]{\fnms{Yuhang}~\snm{Guo}}
\author[A]{\fnms{Kaijun}~\snm{Deng}}
\author[D]{\fnms{Siyang}~\snm{Song}}
\author[A]{\fnms{Jindong}~\snm{Xie}}
\author[A]{\fnms{Wenhui}~\snm{Ma}}
\author[A,B,C]{\fnms{Linlin}~\snm{Shen}\thanks{Corresponding Author. Email: llshen@szu.edu.cn}}

\address[A]{School of Computer Science and Software Engineering, Shenzhen University}
\address[B]{Computer Vision Institute, School of Artificial Intelligence, Shenzhen University, Shenzhen, China}
\address[C]{Guangdong Provincial Key Laboratory of Intelligent Information Processing}
\address[D]{HBUG Lab, University of Exeter}

\begin{abstract}

A key challenge in 3D talking head synthesis lies in the reliance on a long-duration talking head video to train a new model for each target identity from scratch.
Recent methods have attempted to address this issue by extracting general features from audio through pre-training models. However, since audio contains information irrelevant to lip motion, existing approaches typically struggle to map the given audio to realistic lip behaviors in the target face when trained on only a few frames, causing poor lip synchronization and talking head image quality. This paper proposes D$^3$-Talker, a novel approach that constructs a static 3D Gaussian attribute field and employs audio and Facial Motion signals to independently control two distinct Gaussian attribute deformation fields, effectively decoupling the predictions of general and personalized deformations. We design a novel similarity contrastive loss function during pre-training to achieve more thorough decoupling. Furthermore, we integrate a Coarse-to-Fine module to refine the rendered images, alleviating blurriness caused by head movements and enhancing overall image quality. 
Extensive experiments demonstrate that D$^3$-Talker outperforms state-of-the-art methods in both high-fidelity rendering and accurate audio-lip synchronization with limited training data.


\end{abstract}

\end{frontmatter}


\section{Introduction}

Audio-driven Talking Head Synthesis aims to generate consistent and lifelike human talking facial behavior and head movement videos based on the given audio and the target person's portrait. It has been widely applied in fields like film production, game development, online education, and live-streaming sales. Currently, many methods \citep{adnerf,radnerf,dfrf,ernerf,geneface,geneface++,real3d,mimictalk,gaussiantalker,talkinggaussian,gstalker,degstalk,pointtalk,instag} using 3D reconstruction techniques, such as Neural Radiance Fields (NeRF) \citep{nerf} and 3D Gaussian Splatting (3DGS) \citep{3dgs} have been proposed for the talking head synthesis task. These methods use individual speaking videos as the training data to reconstruct the head, which simultaneously employ audio as the driving condition, and finally generate animated talking head portraits. Compared with 2D methods \citep{wav2lip,zhang2021flow,evp,iplap}, they additionally utilize 3D structural information and learn more high-frequency details, thus achieving better detailed textures and image quality. 
However, these 3D-based methods require training a model from scratch for each new identity, necessitating large amounts of talking head video frames to achieve good reconstruction quality and realistic lip movements. A key limitation preventing the real-world application of such approaches is the difficulty of providing long-duration talking videos for training.

Various approaches \citep{dfrf,aenerf,geneface,geneface++,real3d,mimictalk,instag} attempt to address this problem. Some NeRF-based methods \citep{dfrf,aenerf} incorporate features from reference images (i.e., images randomly selected from the training set) for better face reconstruction with limited training data. Some one-shot talking head generation methods \citep{geneface,geneface++,real3d} learn a generic model to avoid dependence on additional video data. However, the former are constrained by the speed of inference and training, while the latter struggle to learn personalized features of the novel identities. Recently, several 3DGS-based methods \citep{talkinggaussian,gaussiantalker,gstalker,degstalk,pointtalk,instag} have been proposed. These methods utilize 3DGS \citep{3dgs} for face reconstruction, learning the deformation of Gaussian attributes through audio. While these methods demonstrate superiority over NeRF-based solutions \citep{adnerf,radnerf,dfrf,ernerf} in terms of synthesis quality and speed under few-shot conditions, it is difficult for them to learn the correct Gaussian attribute deformations only based on audio features when a limited amount of training data is available, making them suffering from poor-quality rendered lip animations.

Based on the above observations, we identify three critical challenges faced by existing few-shot 3D talking head synthesis methods: (1) Imprecise audio-to-lip animation mapping \cite{aenerf}. The limited training data restricts both the accuracy and generalization ability of the mapping from audio to lip movements. It leads to blurry, low-quality, and poorly synchronized lip animations when driving audio is from outside of the training set or from other speakers; (2) The difficulty of decoupling general features from audio. Audio carries identity-specific attributes irrelevant to lip motion, such as pitch and timbre \citep{gstalker}. It is challenging to train a generic model that can decouple lip movement features from the audio of different speakers; and (3) Motion blur caused by head movements. In the 3D talking head synthesis task, the speaker's head movement alters the relative camera pose, potentially causing artifacts during rendering. In this work, we propose D$^3$-Talker to address these issues. Leveraging Audio-Motion Dual-Branch Control Signals, our method adopts Decoupled Deformation Fields to cooperatively represent fine-grained facial motions.

\begin{figure*}[!ht]
\centering
\includegraphics[width=.9\textwidth]{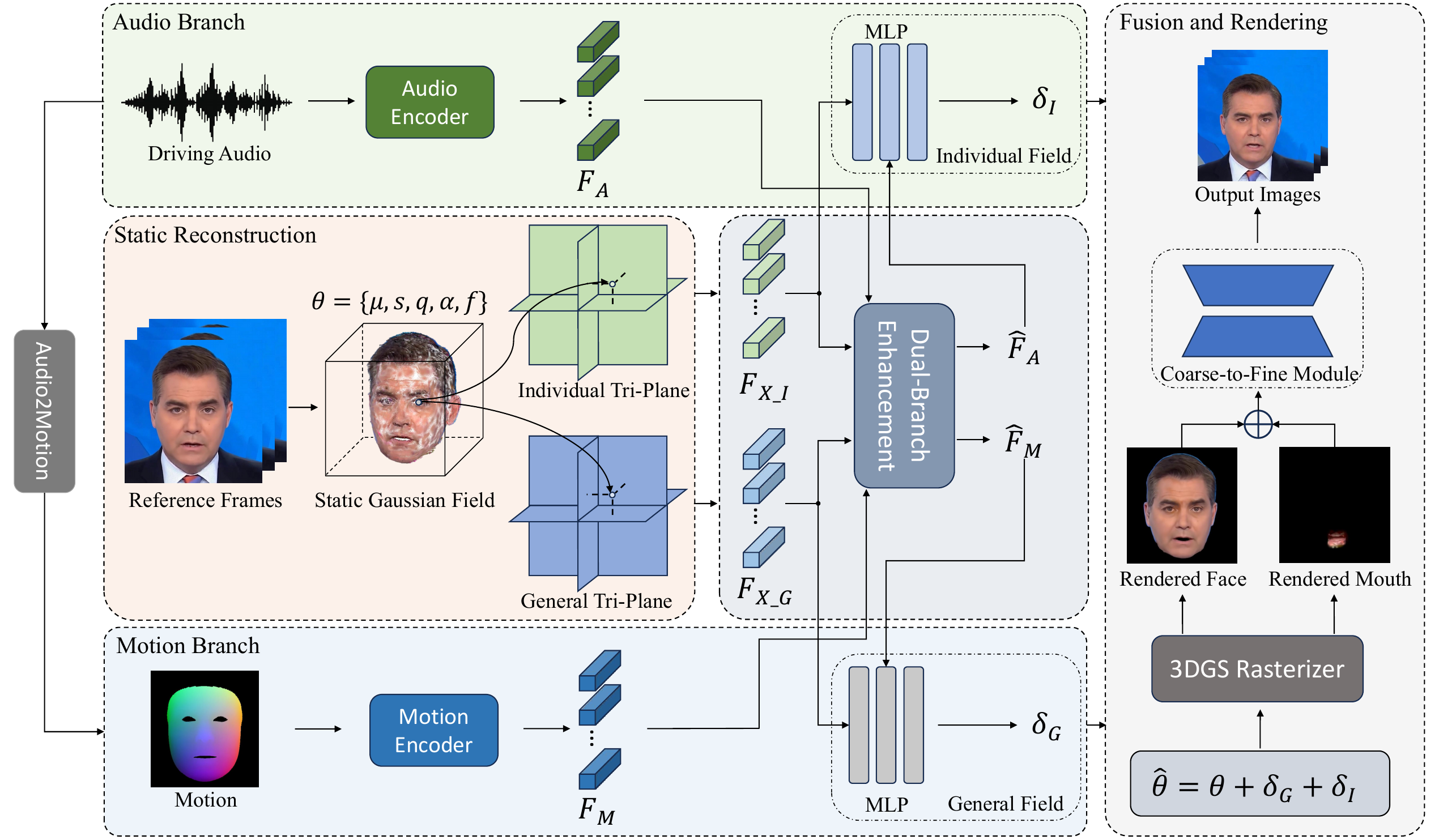}
\captionsetup{justification=raggedright}
\setlength{\belowcaptionskip}{5pt}
\vspace{5pt}
\caption{Overview of the proposed D$^3$-Talker. Given a speech video and the corresponding audio, D$^3$-Talker first builds Static Gaussian Fields. Then the audio branch and motion branch encode audio signal and Facial Motion signal respectively, and the deformations of Gaussian attributes can be co-predicted from the decoupled Individual Field and General Field. After the 3DGS rasterizer renders the 2D images, the Coarse-to-Fine Module refines these coarse images and produce high-quality results.}
\label{fig:overview}
\end{figure*}

Inspired by the identity-free pre-training strategy of InsTaG \citep{instag}, we adopt two decoupled fields to represent the general and personalized 3DGS attributes deformations of the target speaker, respectively. During the two-stage training, we first pre-train a shared Gaussian deformation field on a multi-identity dataset to learn the general deformation features. Then, we adapt it together with a scratch-trained individual deformation field to the target identity to capture additional personal deformation features. However, we found that previous methods \cite{dfrf,aenerf,instag} struggle to train a general base model from multi-identity dataset, as audio contains both speaker's identity information and content information, making it difficult to directly learn the general lip movement feature from audios of different speakers. Therefore, we additionally introduce Facial Motion, a type of 3D face prior feature map extracted from audio by a pre-trained generative model \citep{real3d} to learn general Gaussian deformations, and only employ audio to learn personalized deformations. Furthermore, we design a novel contrastive loss function during the pre-training stage to better decouple speaker-independent features. To confront the challenge posed by motion blur caused by head movements, we introduce the coarse-to-fine module. Specifically, we train a popular neural renderer following existing methods \citep{real3d,gpavatar,gagavatar} to refine the coarse images from Gaussian splatting. 
In summary, this paper presents the following contributions to improve few-shot talking head synthesis:
\begin{itemize}[topsep=0.3cm]
    \item We incorporate an audio-driven Facial Motion prior as a more general prior representation to learn identity-agnostic deformations of Gaussian attributes, proposing a new contrastive loss function to encourage learning generalizability from Facial Motion.
    
    \item We propose a dual-branch signals of audio and Facial Motion to separately control two Gaussian deformation fields, decoupling the prediction of general and individual deformation. 
    
    \item Extensive experiments show that our proposed D$^3$-Talker outperforms the state-of-the-art methods in terms of high-fidelity and accurate audio-lip synchronization with limited training data.
\end{itemize}



\section{Related Work}

\textbf{2D/3D Talking Head Synthesis.} Early methods \citep{2019few,wav2lip} of 2D talking head synthesis directly employ generative models such as GANs \citep{gan} and Auto-encoders \citep{autoencode} to generate talking faces. These methods struggle to achieve high visual fidelity and are incapable of capturing personalized speaking styles. Later, approaches \citep{chen2019,nvp,speech,zhang2021flow,evp,iplap} leverage 2D landmarks or 3D Morphable Models (3DMM) expression coefficients as structural intermediate representations for better face modeling and control. They demonstrate strong generalizability and can be rapidly applied to unseen identities. Nevertheless, the intermediate representation may lead to the loss of high-frequency details. To address the oversight of head structure information, AD-NeRF \citep{adnerf} first introduces NeRF into the talking head synthesis task, as the 3D reconstruction representation of head. Following its successful application, some works \citep{radnerf,ernerf} have realized several improvements in rendering quality and efficiency. Recently, 3DGS introduces an explicit point-based representation for radiance fields and demonstrates higher rendering speed and quality compared to NeRF. Based on this idea, TalkingGaussian \citep{talkinggaussian} and GaussianTalker \citep{gaussiantalker} have implemented talking head reconstruction based on 3DGS. Despite the excellent performance of the above methods, they rely on large-scale datasets to achieve acceptable visual quality.

\textbf{Few-shot Talking Head Synthesis.} Since NeRF and 3DGS require retraining for reconstructing each new scene, their process is not only time-consuming but also highly data-dependent. To alleviate the problem, several few-shot learning methods \citep{grf,pixelnerf,ibrnet,pointnerf,zhao2024chase} have been proposed. Some of them \citep{grf,pixelnerf,ibrnet} have been applied to the talking head synthesis task. DFRF \citep{dfrf} introduces 2D pixel features from reference images (i.e., images randomly selected from the training set) for each 3D query point and reduces training overhead on new identities through a pre-trained base model. AE-Nerf \citep{aenerf} further incorporates audio as guidance for aggregating reference image pixel features and employs two decoupled NeRFs to collaboratively reconstruct the face, thereby improving image quality. MimicTalk \citep{mimictalk} attempts to utilize a one-shot generator, fine-tuning with LoRA \citep{lora} on few minutes videos to capture personalized features. However, significant computational costs during both training and inference stages hinder their practical deployment. InsTaG \citep{instag} utilizes 3DGS for face reconstruction, integrates the training of a universal model, and then adapts to the target individual to learn the personal deformation of Gaussian attributes. It significantly improves the speed of both training and inference. However, it overlooks the difficulty in decoupling general features from audio, which restricts the generalizability to unseen speakers. In this paper, we propose to decouple identity-agnostic features from 3D face prior knowledge to learn a more generalizable deformation field, and use a neural renderer to refine the 3DGS-rendered results. Compared to previous approaches, our method achieves a better balance between rendering speed, image quality, and lip synchronization.


\section{Methodology}

The overview of our D$^3$-Talker is shown in Figure \ref{fig:overview}. D$^3$-Talker starts with static head reconstruction based on Gaussian Splatting \cite{3dgs} (Section \ref{subsec:3_1}). Then, the Audio-Motion Dual-Branch Control Signals (Section \ref{subsec:3_2}) are encoded and separately input into the Decoupled Dual Deformation Fields (Section \ref{subsec:3_3}), co-predicting the deformations of Gaussian attributes. Finally, the results rendered by Gaussian Splatting are refined by the Coarse-to-Fine Module (Section \ref{subsec:3_4}). Additionally, training details are described in Section \ref{subsec:3_5}.

\subsection{Preliminaries}
\label{subsec:3_1}

\textbf{3DGS for Talking Head Synthesis.} 3D Gaussian Splatting (3DGS) \citep{3dgs} represents the scene as a learnable set of Gaussian primitives. Specifically, the $i$-th Gaussian primitive $\mathcal{G}_{i}$ can be described by a set of parameters as:
\begin{equation}
    \theta_{i}=\left\{ \mu_{i},s_{i},q_{i},\alpha_{i},f_{i}\right\},
\label{eq:1}
\end{equation}
where $\mu_{i}\in\mathbb{R}^{3}$ is the center position; $s_{i}\in\mathbb{R}^{3}$ is a scaling factor; $q_{i}\in \mathbb{R}^{4}$ is a rotation quaternion; $\alpha_{i}\in\mathbb{R}^{1}$ denotes the opacity value, and $f_{i} \in \mathbb{R}^{d}$ denotes the $d$-dimensional color feature. By optimizing the parameter $\theta$ for all Gaussians, a static Gaussian field of the head is obtained. Then, we learn the deformation parameters $\delta$ from audio to deform the Gaussian head.

\noindent\textbf{Face-Mouth Region-wise Reconstruction.} Following TalkingGaussian \citep{talkinggaussian} and InsTaG \citep{instag}, we leverage a face-mouth region-wise reconstruction for the head to learn the static fields and deformation fields separately for the face and inside-mouth regions. For the face region, the deformation field predicts the point-wise deformation $\delta_{\text{face}}=\left\{\bigtriangleup \mu,\bigtriangleup s,\bigtriangleup q\right\}$. Then, we use the deformed Gaussian primitive parameters $\theta_{\text{face}}=\left\{ \mu+\bigtriangleup\mu ,s_{i}+\bigtriangleup s,q_{i}+\bigtriangleup q,\alpha,f\right\}$ for rendering. For the inside-mouth region, only the center position deformation $\delta_{\text{mouth}}=\left\{\bigtriangleup \mu\right\}$ is predicted. All deformation fields utilize a tri-plane hash encoder $\mathcal{H}$ \citep{ernerf} to encode the position information of Gaussian primitives, a region attention (RA) \citep{ernerf} module to enhance the spatial perception of audio features, and an MLP decoder to predict the deformation. The process of predicting deformations can be represented as follows:
\begin{equation}
\delta = \mathrm {MLP}\left(\mathcal{H}\left( \mu  \right)\oplus \mathrm {RA}\left( F_{a}  \right )\right),
\label{eq:mlp_deformation}
\end{equation}
where $\mu$ denotes the center position of the query primitive, $F_{a}$ denotes the processed audio features and $\oplus$ denotes concatenation.

\begin{figure}[t]
    \centering 
    \includegraphics[width=0.47\textwidth]{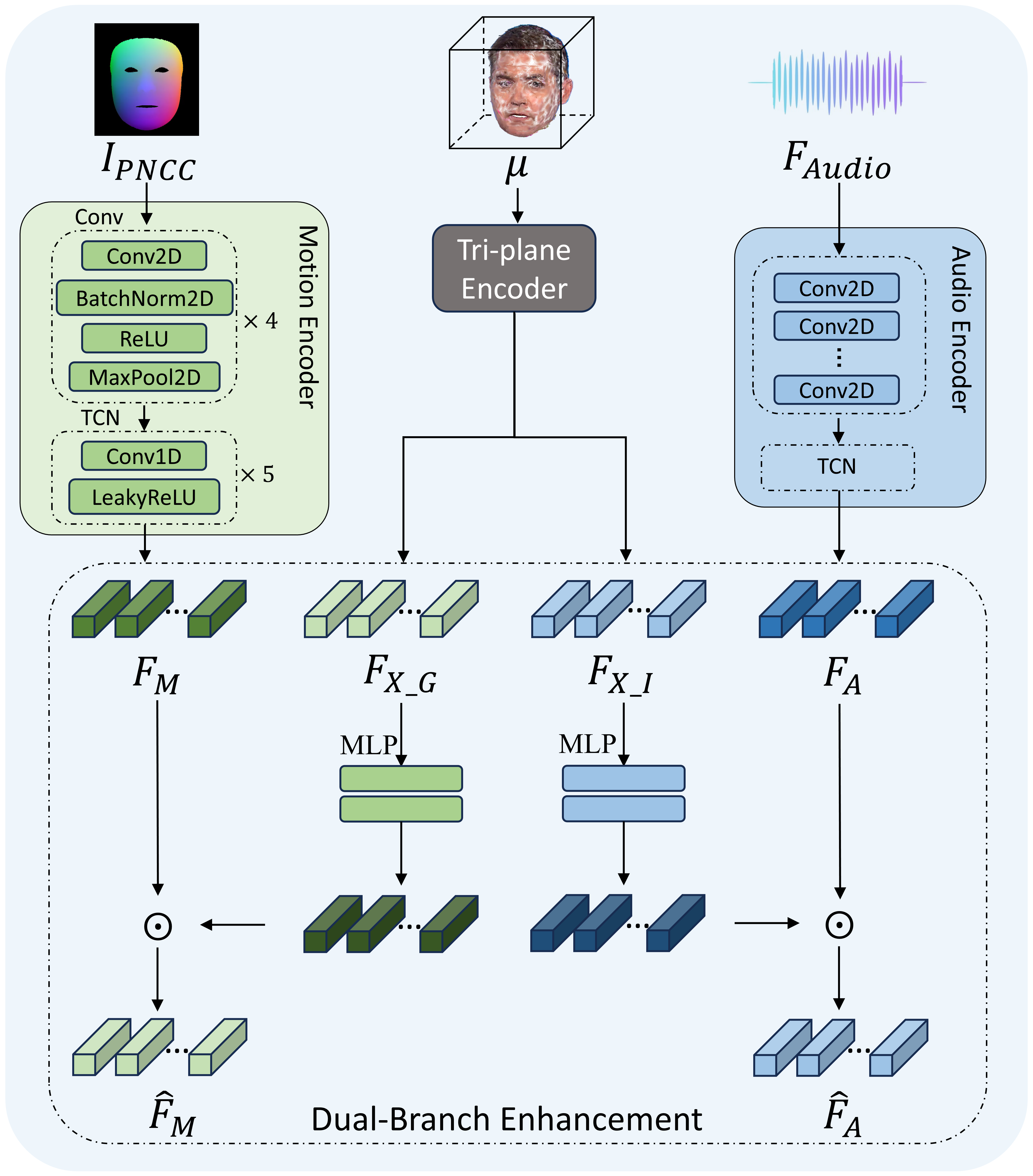} 
    \captionsetup{justification=raggedright}
    \setlength{\belowcaptionskip}{10pt}
    \caption{The detailed process of the dual-branch control signals.}
    \label{fig:dsp} 
\end{figure}


\subsection{Audio-Motion Dual-Branch Control Signals}
\label{subsec:3_2}

We notice that most previous few-shot solutions \cite{dfrf,aenerf,instag} attempt to train a generic model that can extract speaker-invariant lip motion features from  different speakers' audio signals. However, audio signals also carry attributes that are irrelevant to human lip motions and are unique to the speaker, (e.g., pitch and timbre), which prevent these methods from extracting generic features solely from audio. Inspired by PointTalk \citep{pointtalk}, we present Audio-Motion Dual-Branch Control Signals, using Facial Motion together with the audio as the control signal for the deformations of Gaussian attributes, where Facial Motion serves as the general representation of lip motion and audio complement additional individual traits. The detailed process is illustrated in Figure \ref{fig:dsp}.



\noindent\textbf{Facial Motion Signal.} We utilize the pre-trained Audio-to-Motion model from Real3D-Potrait \citep{real3d} to extract motion feature from audio and use PNCC \cite{zhu2016face,kim2018deep} to represent it as $I_{\text{PNCC}}\in\mathbb{R}^{3\times H\times W}$, which is a appearance-agnostic 3D face prior feature map that possesses fine-grained facial expression information based on a 3DMM
face.  
Considering that all Gaussian deformations are based on the static Gaussian field, we design a difference encoder $E_{\text{diff}}$ to better learn subtle facial movements. Specifically, we take the PNCC corresponding to an expression coefficient of $0$ as the canonical representation $I_{\text{PNCC}}^{\text{cano}}$. For the $i$-th frame, the difference is taken between the corresponding motion feature $I_{\text{PNCC}}^{i}$ and $I_{\text{PNCC}}^{\text{cano}}$.
A multi-layer convolution is then utilized to further extract and refine features, mapping them into a corresponding latent space aligned with the audio:
\begin{equation}
\begin{aligned}
F_{\text{Motion}}^{i}=E_{\text{diff}}(\bigtriangleup I_{\text{PNCC}}^{i}) 
=Conv(I_{\text{PNCC}}^{i}-I_{\text{PNCC}}^{\text{cano}}), 
\end{aligned}
\label{eq:pncc_diff_encoder}
\end{equation}
where $F_{\text{Motion}}^{i}\in\mathbb{R}^{1\times32}$. 
Subsequently, we utilize a temporal convolutional network (TCN) to smooth a total of $2j+1$ adjacent frames ($j$ preceding frames and $j$  succeeding frames) for each frame, reducing jitter and eliminating noise signals: 

\begin{equation}
\begin{aligned}
F_{M}^{i}=\mathrm {TCN}(F_{\text{Motion}}^{i-j}\oplus \dots\oplus F_{\text{Motion}}^{i+j}), 
\end{aligned}
\label{eq:tcn_smoothing}
\end{equation}
where $F_{M}^{i}\in\mathbb{R}^{1\times32}$ and $\mathrm {TCN}$ denotes the Temporal Convolutional Network.

\noindent\textbf{Audio Signal.} Following previous audio-driven methods \citep{adnerf,radnerf,dfrf,ernerf,talkinggaussian,instag}, we employ the popular DeepSpeech model \citep{deepspeech} to predict audio features $F_{\text{Audio}}\in\mathbb{R}^{T\times16\times29}$ from the original audio track. To enhance the temporal correlation between adjacent frames, smoothing operation and compression are further carried out to obtain $F_{A}\in\mathbb{R}^{T\times32}$.


\noindent\textbf{Dual-Branch Enhancement.} To capture the relationships between two types of signal features and different spatial regions, we propose the Dual-Branch Enhancement module. For the motion branch, we employ the tri-plane hash encoder $\mathcal{H}$ \citep{ernerf} to extract multi-resolution regional information $F_{X\_G}$ from the center position of static Gaussians $\mu$. Then, we feed $F_{X\_G}$ to a two-layer MLP to generate the region attention vector describing the face spatial information. Finally, the region-aware feature $\hat{F}_M$ is calculated through the Hadamard product of $F_M$ and the region attention vector as: 
\begin{equation}
\hat{F}_M = \mathrm {MLP}\left(\mathcal{H}\left(\mu \right)\right)\odot F_M,
\label{eq:dual_branch_enhancement}
\end{equation}
where $\odot$ denotes Hadamard product. The processing of the audio branch is done in the same way.

\subsection{Decoupled Dual Deformation Fields}
\label{subsec:3_3}

Based on the Dual-Branch Control Signals from Section \ref{subsec:3_2}, we introduce the Decoupled Dual Deformation Fields and General Field and Individual Field, to separately predict the general and personalized deformations of Gaussian attributes. Furthermore, a two-stage training strategy is employed to train the Decoupled Dual Deformation Fields, during which we also design a novel Similarity Contrastive Loss to encourage General Field to be generalizable.



\noindent\textbf{General Field.} General Field aims to predict identity-agnostic deformations of Gaussian attributes. To avoid the influence of information irrelevant to lip motion on the General Field, we use Facial Motion instead of audio as the input to the General Field. Given the center position $\mu$, the point-wise deformation parameter $\delta_G$ is predicted with the condition feature $\hat{F}_M$: 
\begin{equation}
\delta_G = \mathrm {MLP}\left(\mathcal{H}\left( \mu  \right)\oplus \hat{F}_M\right).
\label{eq:general_field}
\end{equation}

\noindent \textbf{Individual Field.} To learn more personalized and expressive facial movements, we employ an additional deformation field, termed Individual Field, for each identity. The Gaussian deformations for each identity are jointly learned by its corresponding Individual Field and pre-trained General Field. Since audio contains identity-related information, we leverage audio to learn personalized deformations $\delta_ I$. For the $k$-th identity, a process similar to the General Field is applied to predict deformations:
\begin{equation}
\delta_I^k = \mathrm {MLP}\left(\mathcal{H}^k\left( \mu^k\  \right)\oplus \hat{F}^k_A\right),
\label{eq:individual_field}
\end{equation}
where $k$ represents the identity order. $\delta_I^k$, $\mathcal{H}^k$ and $\hat{F}^k_A$ denote the personalized deformation, Hash encoder and region-aware audio feature for the $k$-th individual, respectively.

\noindent \textbf{Two-stage Training Strategy.} A two-stage training strategy that is similar to previous works \cite{dfrf,aenerf,instag} is adopted, allowing prior learning of common motion knowledge from long videos as compensation for the target identity. The first stage, the pre-training stage, trains the shared General Field on a multi-identity dataset to learn the general deformation features. 

In the second stage, the adaptation stage, given a short video clip of the target identity, we utilize the pre-trained General Field, together with a scratch-trained Individual Field, to adapt personalized features of the target identity. We posit that, during adaptation to a target identity, personalized information is more readily derived from audio. Therefore, unlike in the adaptation of InsTaG \citep{instag}, our Individual Field predicts the same deformation attributes as the General Field, and the two are directly added to deform the static field of the target identity:
\begin{equation}
\tilde{\theta}  = \theta + \delta , \quad \delta  = \delta_G + \delta_I,
\label{eq:3}
\end{equation}
where $\theta $ is the static Gaussian attributes. After that, a 2D image $I$ can be rendered as:
\begin{equation}
I  = \mathcal{R}\left(\tilde{\theta}, \ \left[R, t\right]\right) ,
\label{eq:3}
\end{equation}
where $\mathcal{R}(\cdot)$ is the 3DGS rasterizer and $[R,t]$ represents the camera pose.

\noindent \textbf{Similarity Contrastive Loss.} To better decouple personalized and general deformations, InsTaG \citep{instag} introduced the Negative Contrast Loss $\mathcal{L}_C$ to encourage the diversity of personalized deformations. However, $\mathcal{L}_C$ only focuses on discriminating personalized deformations across Individual Fields while neglecting the critical role of the General Field in the pre-training stage. To address this limitation, we introduce a novel Similarity Contrastive Loss for each identity that explicitly enhances the similarity between features learned by the General Field and those derived from the corresponding audio of that identity. 

Given a total of $N$ identities during pre-training, there exists a General Field and $N$ identity-dependent Individual Fields. For the $k$-th identity, our objective is to enhance the similarity between the universal $\mu$-deformation $\bigtriangleup\mu_{G}$ from the General Field and the personalized $\mu$-deformation $\bigtriangleup\mu_{I}^{k}$ from the $k$-th Individual Field. Meanwhile, the similarity between $\bigtriangleup\mu_{G}$ and $\mu$-deformation from other Individual Fields queried by the same audio feature $\hat{F}^k_A$ should be minimized. Therefore, the loss function $\mathcal{L}_{SC} \left(k\right)$ is constructed as:
\begin{equation}
\mathcal{L}_{SC} \left(k\right) = -\mathrm{log}\frac{\mathrm{exp}\left(sim(\bigtriangleup\mu_{G}, \bigtriangleup\mu_{I}^{k})/\tau \right)}{\sum_{i=1}^{N}\mathrm{exp}\left( sim(\bigtriangleup\mu_{G}, \bigtriangleup\mu_{I}^{i})/\tau\right)},
\label{eq:sc_loss}
\end{equation}
where $sim(\cdot , \cdot )$ denotes the cosine similarity function, $\tau$ is the temperature factor.

\begin{table*}[!htbp]
\centering
\caption{Quantitative comparison under the self-driven setting with 10s training data.}
\begin{tabular}{l|c|ccc|ccc|cc}
\hline
\multirow{3}{*}{Method} & \multirow{4}{*}{Type}     & \multicolumn{3}{c|}{Visual Quality}                                       & \multicolumn{3}{c|}{Lip Synchronization}                                           & \multicolumn{2}{c}{Efficiency}     \\
                        & & \multirow{2}{*}{PSNR↑} & \multirow{2}{*}{SSIM↑} & \multirow{2}{*}{LPIPS↓} & \multirow{2}{*}{LMD↓} & \multirow{2}{*}{Sync-C↑} & \multirow{2}{*}{Sync-D↓ } & Training  & \multirow{2}{*}{FPS↑}  \\
                       & &                        &                        &                         &                       &                             &                              &Time ↓     &                       \\ 
Ground Truth            & &  N/A                   & 1                      & 0                       & 0                     & 8.685                       &  6.852                       & -         &  -                     \\ \hline
RAD-NeRF \cite{radnerf} & \multirow{3}{*}{NeRF-based} & 29.511                 & 0.907                  & 0.053                   & 3.678                 & 1.792                       & 12.575                      & 6 hours   & 24                    \\
DFRF \cite{dfrf} \mysmall{ECCV'22}       & & 30.413                 & 0.909                  & 0.060                   & 3.460                 & 3.658                       & 10.927                       & 9 hours   & 0.02                     \\
ER-NeRF \cite{ernerf} \mysmall{ICCV'23} & & 30.966                 & 0.913                  &\underline{0.041}        & 3.656                 & 2.363                       & 12.108                       & 2 hours   & 30                       \\ \hline
GaussianTalker  \cite{gaussiantalker} \mysmall{ACM MM'24}         & \multirow{4}{*}{3DGS-based} & 30.895                 & \underline{0.921}      & \textbf{0.039}          & 3.793                 & 2.086                       & 12.523                       & 53 min    & 62                          \\
TalkingGaussian \cite{talkinggaussian} \mysmall{ECCV'24}       & & 30.833                 & 0.918                  & \textbf{0.039}          & 3.473                 & 3.055                       & 11.553                       & 37 min    & \textbf{89}                       \\
InsTaG \cite{instag} \mysmall{CVPR'25}   &  & \underline{31.129}     & \underline{0.921}      & 0.042                   & \underline{3.195}     & \underline{4.411}           & \underline{10.379}           & \textbf{16 min}  & \underline{69}                            \\ 
D$^3$-Talker(Ours)     &  & \textbf{31.518}        & \textbf{0.927}         & 0.043                   & \textbf{3.194}        & \textbf{6.015}              & \textbf{9.037}               & \underline{32 min}   &  65                      \\ \hline
\end{tabular}
\label{tab:10s}
\vspace{0.3cm}
\end{table*}

\begin{table*}[]
\centering
\caption{Quantitative comparison under the self-driven setting with different training data amounts.}
\centering

\begin{tabular}{c|cccc|cccc|cccc}
\hline
\multicolumn{1}{c|}{\multirow{2}{*}{Method}} & \multicolumn{4}{c|}{DFRF \citep{dfrf}} & \multicolumn{4}{c|}{InsTaG \citep{instag}}  & \multicolumn{4}{c}{D$^3$-Talker}        \\
\multicolumn{1}{c|}{}                        & 5s     & 10s     & 15s    & 20s     & 5s     & 10s     & 15s     & 20s               & 5s              & 10s             & 15s             & 20s   \\ \hline\hline
PSNR↑                                        & 29.774 & 30.413  & 30.772 & 30.896  & 30.493 & 31.129  & 31.686  & 32.068            & \textbf{30.595} & \textbf{31.518} & \textbf{32.291} & \textbf{32.566} \\[0pt]
SSIM↑                                        & 0.902  & 0.909   & 0.913  & 0.915   & \textbf{0.916}  & 0.921   & 0.928   & 0.932             & \textbf{0.916}  & \textbf{0.927}  & \textbf{0.934}  & \textbf{0.936}  \\[0pt]
LPIPS↓                                       & 0.063  & 0.060   & 0.059  & 0.059  & \textbf{0.047} & \textbf{0.042} & \textbf{0.040} & \textbf{0.038}    & 0.050           & 0.043           & 0.043      & 0.042           \\[0pt]
LMD↓                                         & 3.691  & 3.460   & 3.341  & 3.290   & 3.395  & 3.195   & 3.182   & 3.140             & \textbf{3.283}  & \textbf{3.194}  & \textbf{3.156}  & \textbf{3.060}  \\[0pt]
Sync-C↑                                      & 2.775  & 3.658   & 4.040  & 4.329   & 3.992  & 4.411   & 4.888   & 4.984             & \textbf{5.761}  & \textbf{6.015}  & \textbf{6.538}  & \textbf{6.453}  \\[0pt]
Sync-D↓                                      & 11.757 & 10.927  & 10.588 & 10.360  & 10.661 & 10.379  & 10.023  & 9.900             & \textbf{9.224}  & \textbf{9.037}  & \textbf{8.659}  & \textbf{8.659}  \\ \hline
\end{tabular}
\label{tab:diffsec}
\vspace{0.3cm}
\end{table*}

\subsection{Coarse-to-Fine Module}
\label{subsec:3_4}

Previous methods \citep{talkinggaussian,instag} simply combine the rendered results from the mouth and face regions to produce final output images. However, due to limited camera viewpoints in fewer training image frames, the 3DGS rasterizer tends to render artifacts, especially when the speaker's head moves. To this end, we propose the Coarse-to-Fine module to refine the coarse images from Gaussian Splatting. Specifically, we first sum the results rendered from the two regions. Then, we feed them into a StyleUnet-based neural renderer, following existing methods \citep{real3d,gpavatar,gagavatar}. We treat the renderer as an inpainting model rather than super-resolution, with both input and output resolutions being $512\times 512$. During training, we first train a base model from scratch on the multi-identity datasets from the pre-training stage, and then fine-tune it on the target identity.

\subsection{Stage-wise Training Details}
\label{subsec:3_5} 

\noindent\textbf{Pre-training.} In the first stage, the pre-training stage, we collect videos of $N$ distinct identities to form the training dataset and initialize the static Gaussian fields for each identity. Following the original 3DGS \citep{3dgs}, we utilize a combination of L1 pixel-wise loss $\mathcal{L}_1$ and a D-SSIM \citep{dssim} term $\mathcal{L}_{\text{D-SSIM}}$. After that, we add the shared General Field and $N$ identity-dependent Individual Fields to predict the deformation parameters and additionally use the Similarity Contrastive Loss $\mathcal{L}_{\text{SC}}$ to supervise this process. We also train our Coarse-to-Fine model from scratch using the coarse images rendered during pre-training concurrently. Our primary objective during training is to ensure that the reenacted fine image aligns with the target image. We employ L1 loss and perceptual loss \citep{perceptual,lpips} for this purpose:
\begin{equation}
\mathcal{L}_{\text{C2F}} = \left \| I_f - I_t\right \| +\lambda_p \left \| \varphi (I_f) - \varphi (I_t)\right \|,
\label{eq:3}
\end{equation}
where $I_f$ is the generated fine image, $I_t$ is the target image, and $\varphi$ is the AlexNet \citep{imagenet} used in the perceptual loss. The overall loss function for pre-training is:
\begin{equation}
\mathcal{L}_{\text{pre}} = \mathcal{L}_1 + \lambda_{\text{D-SSIM}}\mathcal{L}_{\text{D-SSIM}} + \lambda_{\text{SC}}\mathcal{L}_{\text{SC}} + \lambda_{\text{C2F}} \mathcal{L}_{\text{C2F}}.
\label{eq:3}
\end{equation}


\noindent\textbf{Adaptation.} During the adaptation stage, we jointly train an initialized Individual Field with the pre-trained General Field and Coarse-to-Fine model. Following InsTaG \citep{instag}, we add the geometry loss $\mathcal{L}_{\text{Geo}}$ to regularize the depth and surface normals. The overall loss function can be constructed as:
\begin{equation}
\mathcal{L}_{\text{ada}} = \mathcal{L}_1 + \lambda_{\text{D-SSIM}}\mathcal{L}_{\text{D-SSIM}} + \lambda_{C}\mathcal{L}_{C} + \lambda_{\text{Geo}}\mathcal{L}_{\text{Geo}}.
\label{eq:3}
\end{equation}


\begin{figure*}[!ht]
\centering
\includegraphics[width=.95\textwidth]{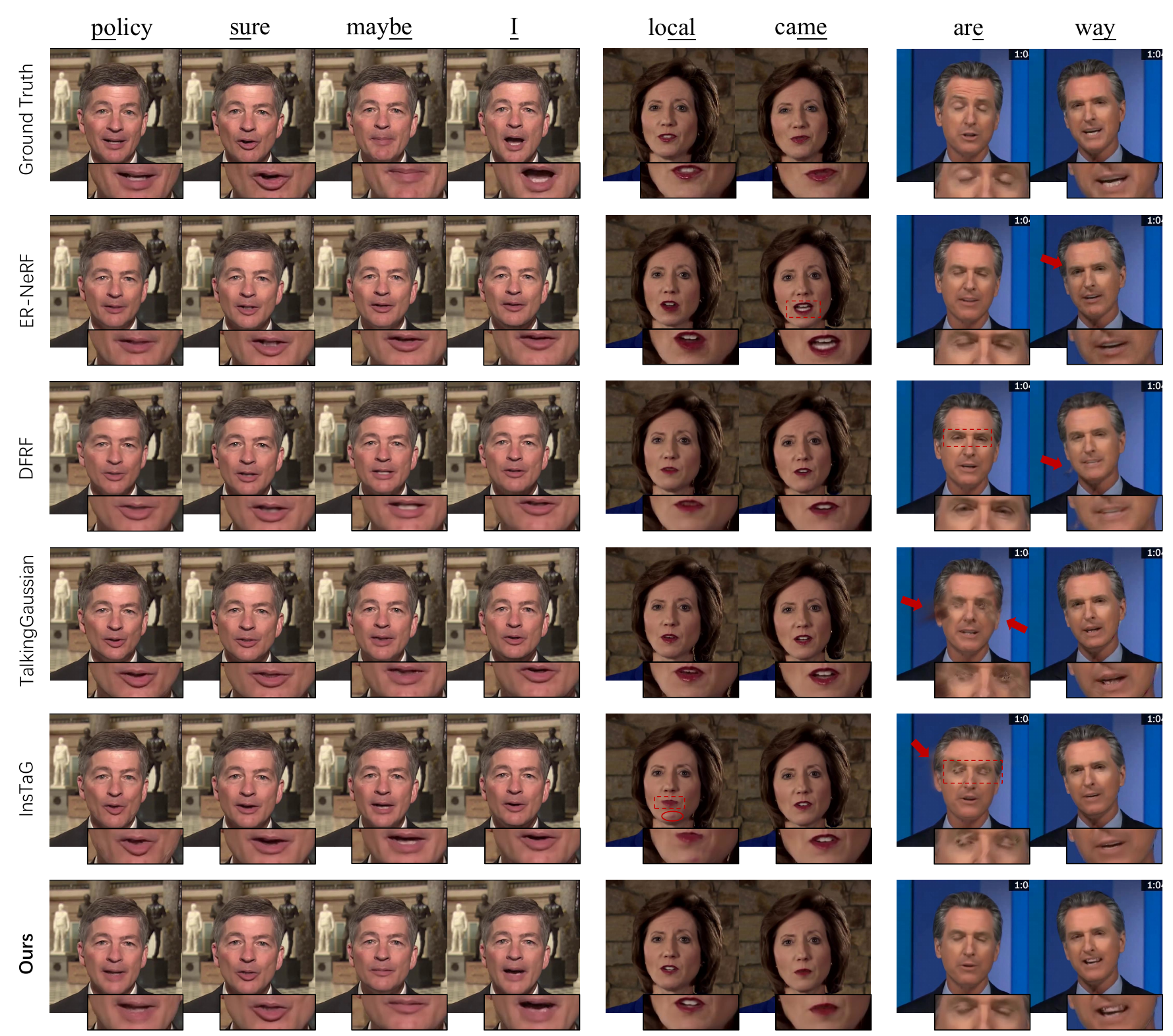}
\captionsetup{justification=raggedright}
\setlength{\belowcaptionskip}{10pt}
\caption{Qualitative comparison of generated key frames under the self-driven setting with different methods. The results depict the lip shape conforming to specific phonemes in the spoken words ‘policy’, ‘sure’, ‘maybe’, ‘I’, ‘local’, ‘came’, ‘are’ and ‘way’.}
\vspace{-3pt}
\label{fig:qualitative1}
\end{figure*}

\section{Experiments}
\subsection{Experiment Setting}
\noindent\textbf{Datasets.} Following existing practices \citep{dfrf,talkinggaussian,instag}, we collect 6 speaking videos provided by AD-NeRF \citep{adnerf}, DFRF \citep{dfrf} and HDTF \citep{zhang2021flow}, including English and Chinese languages, for testing. For pre-training, we use 5 long speaking videos of different identities from ER-NeRF \citep{ernerf} and Geneface \citep{geneface} to train the General Field and the Coarse-to-Fine model. There is no overlap between the training set and the test set. Each raw video is resampled to 25 FPS and the resolution is resized as $512\times512$. 

\noindent\textbf{Comparison Baselines.} We compare our method with three NeRF-based methods RAD-NeRF \citep{radnerf}, DFRF \citep{dfrf} and ER-NeRF \citep{ernerf}, three 3DGS-based methods GaussianTalker \citep{gaussiantalker}, TalkingGaussian \citep{talkinggaussian} and InsTaG \citep{instag}. For each method, we use their official code, and a DeepSpeech model \citep{deepspeech} is used to extract basic audio features. DFRF and InsTaG also have the base model like our method, and we use their official pre-trained models.

\noindent\textbf{Evaluation Metrics.} We employ 6 evaluation metrics commonly used in talking head synthesis to comprehensively evaluate the visual quality and lip synchronization. Specifically, we utilize Peak Signal-to-Noise Ratio (PSNR) and Structural Similarity Index Metric (SSIM) \citep{dssim} to evaluate the image level quality of generated results, Learned Perceptual Image Patch Similarity (LPIPS) \citep{lpips} to evaluate the feature level quality. For lip synchronization, we use Landmark Distance (LMD) \citep{lmd} which quantifies the distance between lip landmarks, SyncNet \citep{syncnet} confidence (Sync-C) and distance (Sync-D) to assess the synchronization between lip movements and audio.

\subsection{Quantitative Evaluation}
\noindent\textbf{Results under Self-driven Setting.} The evaluation results of self-driven setting with 10s training data are illustrated in Table \ref{tab:10s}. It can be seen that our method achieves the best performance in multiple aspects compared to other methods. Specifically, in terms of image quality, our method outperforms other methods in both PSNR and SSIM. Particularly, our method demonstrates significantly better performance in lip synchronization metrics compared to all other methods. This is mainly due to the assistance of the Facial Motion. Additionally, our method maintains excellent performance in both training time and inference FPS, both of which are close to InsTaG. While our method does not achieve the best efficiency, all baselines can not simultaneously ensure high synchronization and real-time inference, in which our method strikes a good balance.

To more comprehensively evaluate the performance of our method in practical applications, we compare quantitative metrics under varying training data amounts with two few-shot methods, DFRF and InsTaG. The results with training data lengths of 5s, 10s, 15s and 20s are shown in Table \ref{tab:diffsec}. It is evident that our method still surpasses other methods in most aspects, maintaining high image generation quality and audio visual alignment.

\begin{table}[t]
\centering
\caption{Quantitative comparison under the cross-driven setting with 10s training data.}
\begin{tabular}{lcccc}
\hline
\multirow{2}{*}{Method} & \multicolumn{2}{c}{Audio A (Obama)} & \multicolumn{2}{c}{Audio B (May)} \\ \cline{2-5} 
                        & Sync-C↑          & Sync-D↓          & Sync-C↑         & Sync-D↓         \\
Ground Truth            & 8.456            & 6.841            & 8.809           & 5.757           \\ \hline
RAD-NeRF \citep{radnerf} & 3.031           & 10.986           &  3.399          &  10.814         \\
DFRF  \citep{dfrf}      & 3.256            & 11.056           & 4.196           & 10.390          \\
ER-NeRF \citep{ernerf}   & 2.637            & 11.892           & 2.909           & 11.923          \\
GaussianTalker \citep{gaussiantalker}  & 1.860            & 13.227           & 1.501           & 13.558          \\
TalkingGaussian \citep{talkinggaussian}    & 3.810            & 10.519           & 3.291           & 11.284          \\
InsTaG  \citep{instag}    & 3.510            & 11.275           & 4.484           & 10.876          \\ \hline
D$^3$-Talker            & \textbf{6.966}   & \textbf{8.345}   & \textbf{6.142}  & \textbf{9.262}   \\ \hline
\end{tabular}
\vspace{-5pt}
\label{tab:crossID}
\end{table}

\noindent\textbf{Results under Cross-driven Setting.}
For the cross-driven setting, we extract one unseen audio clip from Obama's speaking video \citep{adnerf} and one from May \citep{geneface} to drive each method and compare lip synchronization. Owing to the absence of ground truth, we only evaluate the accuracy of lip synchronization using SyncNet confidence score (Sync-C) and distance score (Sync-D). We consider the metrics measured from the original video as the ground truth. The results are presented in Table \ref{tab:crossID}, which shows that our D$^3$-Talker offer broader application value and robustness across various scenarios.

\subsection{Qualitative Evaluation}
\begin{figure}[t]
    \centering 
    \includegraphics[width=0.48\textwidth]{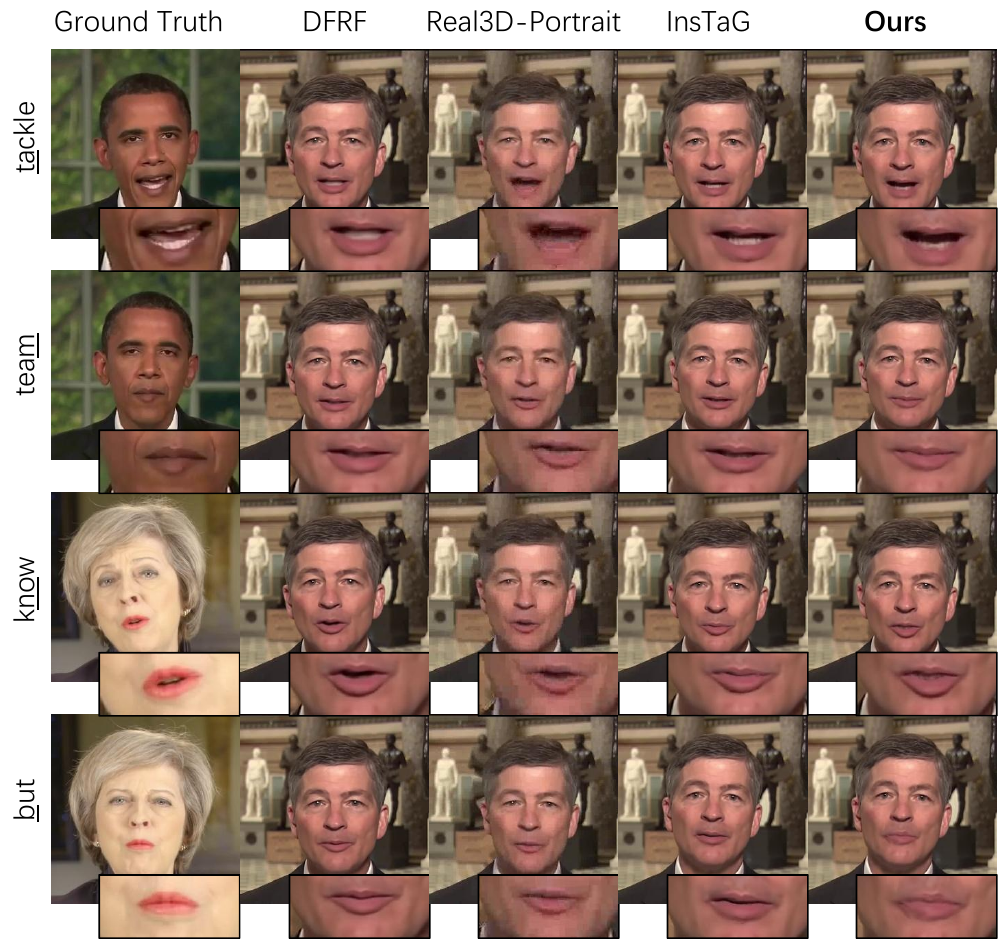} 
    \captionsetup{justification=raggedright}
    \setlength{\belowcaptionskip}{10pt}
    \caption{Qualitative comparison of generated key frames under the cross-driven setting with different methods. }
    \vspace{5pt}
    \label{fig:qualitative_cross} 
\end{figure}

To more intuitively evaluate the image quality and lip synchronization, we showcase key frames and close-up details of generated videos from self-driven and cross-driven settings. The key frames of three portraits under self-driven setting with 10 seconds training data are displayed in Figure \ref{fig:qualitative1}. We label the specific word and phoneme corresponding to each frame above the image. It can be observed from the four frames of the first identity that our method behaves better in lip synchronization, accurately capturing various phonetic elements. Particularly, for the frame corresponding to the spoken word ‘maybe’, our method is the only one that achieves lip closure. As for the word ‘I’, we also exhibit the most realistic mouth opening, closely matching the style of the ground truth. Furthermore, while other methods generate blurred or severely detail loss in the reconstruction of face and eye blinking for the second and third portrait (red mark), our method generates photorealistic images with delicate details in non-rigid regions like eyes and wrinkles, benefiting from the Coarse-to-Fine module. The visualization results under the cross-driven setting are shown in Figure \ref{fig:qualitative_cross}. These results demonstrate that our method maintains impressive lip synchronization capability even with cross-identity and cross-gender audio inputs.

\begin{table}[t]
\centering
\caption{Ablation study on different settings.}
\setlength{\tabcolsep}{6pt}
\begin{tabular}{lcccc}
\hline
Setting            & PSNR↑            & SSIM↑          & LMD↓           & Sync-C↑   \\  
Ground Truth      & N/A              & 1               & 0              & 8.685    \\ \hline
D$^3$-Talker      & \textbf{32.566}  & 0.936  & \textbf{3.060} & \textbf{6.453}    \\
w/o FM            & 32.176           & 0.932           & 3.251          & 5.012     \\
w/o SCLoss        & 32.539           & 0.936  & 3.228          & 6.021    \\
w/o C2F           & 32.049           & 0.931           & 3.126          & 6.333     \\  \hline
Audio Only        & 32.176          & 0.932            & 3.251          & 5.012    \\
Facial Motion Only  & 32.552          &\textbf{0.938}    & 3.133          & 6.128   \\ \hline 
General Field Only    & 30.546          & 0.913           & 3.472          & 4.414   \\
Individual Field Only & 30.391          & 0.901           & 3.930          & 0.681 \\ \hline
NCLoss (InsTaG) Only  & 32.539           & 0.936       & 3.228          & 6.021   \\ 
SCLoss (Ours) Only    & 32.528          & 0.935        & 3.157         & 6.217   \\ 
None                  & 30.584          & 0.916           & 3.907          & 5.004  \\ \hline 
$\Delta \mu$          & 32.356          & 0.935          & 3.114          & 6.326  \\
$\Delta \mu,\!\Delta r,\!\Delta s,\!\Delta \alpha,\!\Delta SH$ & 32.552      & 0.935            & 3.117           & 5.495          \\
$\Delta \mu,\!\Delta r,\!\Delta s$      & \textbf{32.566} & 0.936     & \textbf{3.060} & \textbf{6.453}       \\ \hline 
\end{tabular}
\label{tab:ablation1}
\vspace{0.0cm}
\end{table}

\subsection{Ablation Study}
In this section, we provide multi-faceted ablation studies to validate the effectiveness of our contributions and design choices. 

\begin{figure}[t]
    \centering 
    \includegraphics[width=0.47\textwidth]{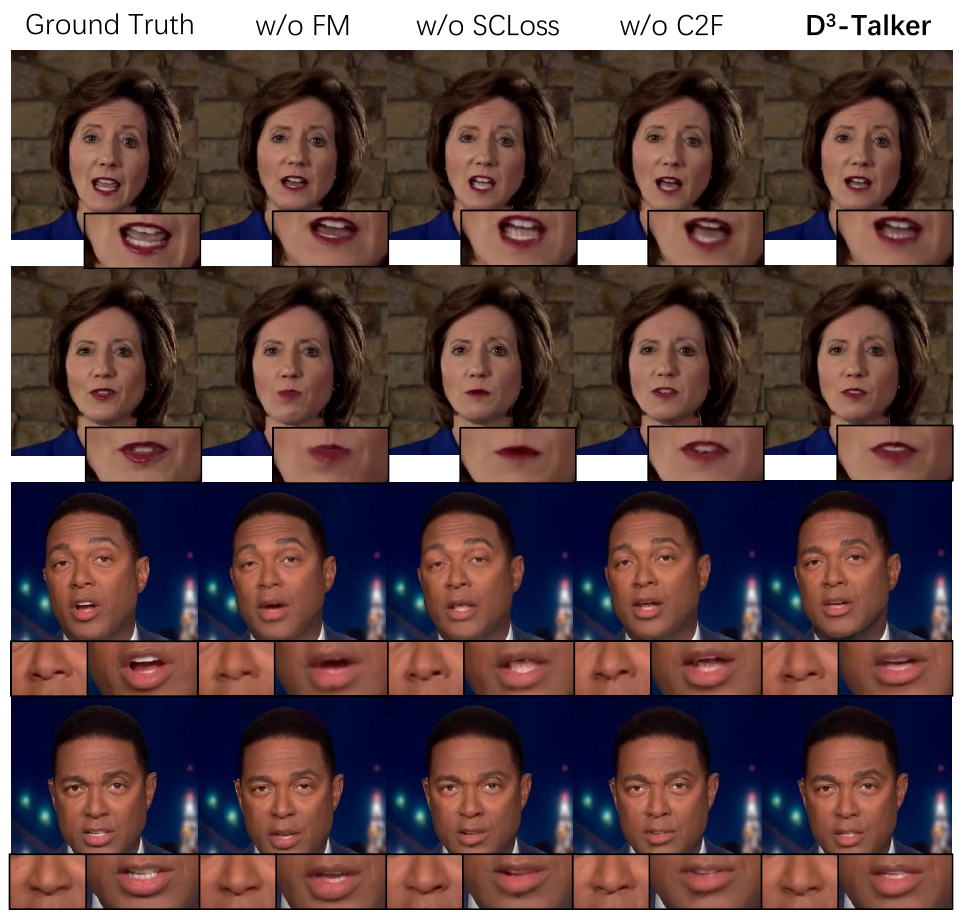} 
    \captionsetup{justification=raggedright}
    \setlength{\belowcaptionskip}{10pt}
    \caption{Qualitative results of the ablation study.}
    \vspace{5pt}
    \label{fig:qualitative2} 
\end{figure}

\noindent\textbf{Main Contributions of Our Work.} Experiments in Table \ref{tab:ablation1} (lines 3-6) show that removing Facial Motion (FM) or Similarity Contrastive Loss (SCLoss) results in lower quality in lip synchronization, while the output images without the Coarse-to-Fine (C2F) module’s refinement produce relatively poor visual quality. We also visualize the results with 10s training data in Figure \ref{fig:qualitative2} for intuitive comparison, demonstrating the effectiveness of our contributions. 

\noindent\textbf{Dual-Branch Control Signals.} To evaluate the necessity of our Dual-Branch Control Signals, we respectively use only audio or only Facial Motion as the control signals for all deformation fields. The results in Table \ref{tab:ablation1} (lines 7-8) show that both control signals contain specific lip-motion features and combining these two branches achieves the best performance among all settings. 

\noindent\textbf{Decoupled Deformation Fields.} To validate the effectiveness of the two deformation fields in our method, we use the deformations from General Field or Individual Field solely to deform the Gaussian head in Table \ref{tab:ablation1} (lines 9-10).  Both deformation fields are valid, and General Field can more effectively influence lip synchronization.

\noindent\textbf{Contrastive Loss.} We compare our Similarity Contrastive Loss (SCLoss) with the Negative Contrast loss (NCLoss) of InsTaG. The results in Table \ref{tab:ablation1} (lines 11-13) indicate the superiority of our SCLoss over NCLoss. 

\noindent\textbf{Selection of Deformed Attributes.} We investigate different selections of Gaussian attributes for deformation in the Individual Field in Table \ref{tab:ablation1} (lines 14-16). Only controlling the center positions leads to loss of reconstruction accuracy. However, deformation of all Gaussian attributes results in a lower performance in lip synchronization.

\vspace{-0.2cm}
\section{Conclusion}
In this paper, we present D$^3$-Talker, a 3D-Gaussian based method for few-shot talking head synthesis that addresses key limitations in existing approaches. By incorporating an audio-driven Facial Motion prior, our method effectively learns identity-agnostic general representations. The dual-branch architecture, which uses audio and Facial Motion signals to independently control two Gaussian attribute deformation fields, successfully decouples general and individual deformation predictions. Our comprehensive experiments demonstrate that D$^3$-Talker achieves superior image quality and lip synchronization performance with limited training data, outperforming state-of-the-art methods across various evaluation metrics and scenarios.


\begin{ack}
This work was funded by Guangdong Basic and Applied Basic Research Foundation under Grant 2023A1515010688, and Guangdong Provincial Key Laboratory under Grant 2023B1212060076.
\end{ack}

\vspace{-0.1cm}

\bibliography{mybibfile}

\end{document}